\pdfoutput=1

\RequirePackage[hyphens]{url}
\documentclass[11pt]{article}

\usepackage[final]{acl}

\usepackage{times}
\usepackage{latexsym}

\usepackage[T1]{fontenc}

\usepackage[utf8]{inputenc}

\usepackage{microtype}

\usepackage{inconsolata}

\usepackage{graphicx}
\usepackage{subcaption}

\usepackage{amsmath}
\usepackage{amssymb}

\usepackage{cleveref}
\usepackage{xspace} 

\usepackage{tabularx}
\usepackage{listings}
\usepackage{array}
\usepackage{multirow}
\usepackage{float} 

\PassOptionsToPackage{hyphens}{url}\usepackage{hyperref}

\definecolor{vova}{rgb}{0.1, 0.6, 0.2}

\newcommand{\metric}{\textsc{MATCH}\xspace}

\newcommand{\codebertscore}{{CodeBERTScore}\xspace}
\newcommand{\icescore}{{ICE-Score}\xspace}
\newcommand{\codescore}{{CodeScore}\xspace}

\title{\metric: Task-Driven Code Evaluation through Contrastive
Learning}
\author{
Marah Ghoummaid, Vladimir Tchuiev, Ofek Glick,\\ \textbf{Michal Moshkovitz, \and Dotan Di Castro} \\
Bosch Research \\
\texttt{marahghoummaid@gmail.com, vladimir.tchuiev@il.bosch.com, } \\
\texttt{ofek.gluck@gmail.com,michal.moshkovitz@il.bosch.com, }
\\
\texttt{ dotan.dicastro@il.bosch.com}
}

\begin{document}
\maketitle
\begin{figure*}[h]
  \includegraphics[width=0.85\textwidth]{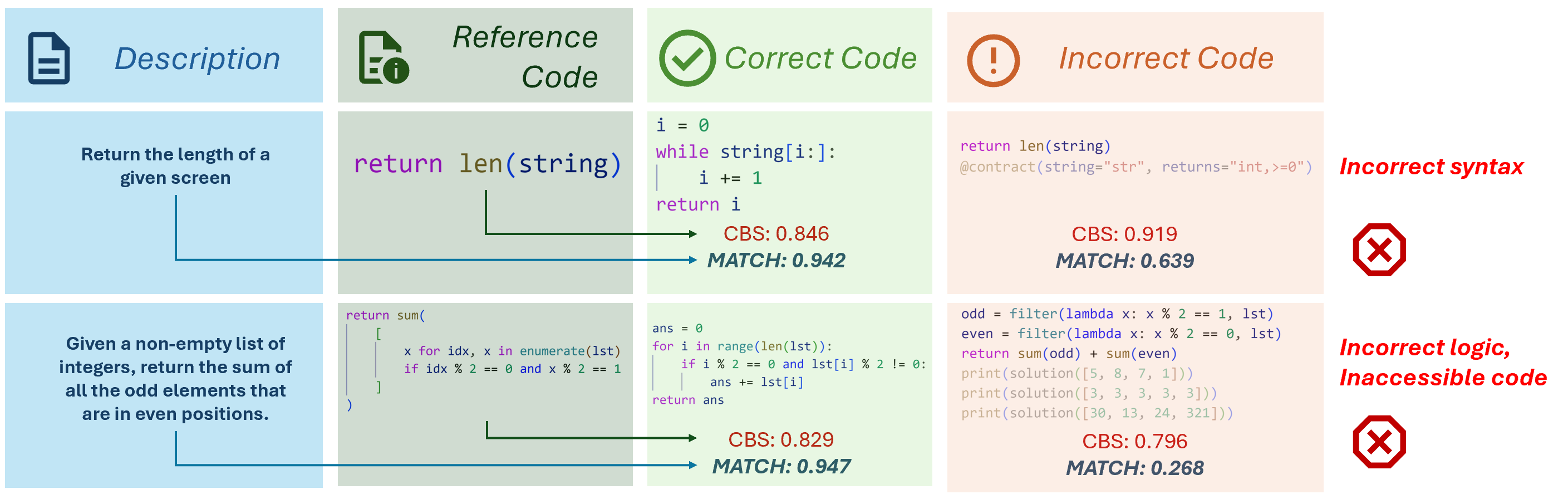}
  \centering
  \caption{Two examples from the HumanEval dataset that illustrate \metric's evaluation ability for correct and incorrect code snippets given a description (for \metric) and a reference code (for CodeBERTScore, denoted as \emph{CBS}). The type of incorrectness for the incorrect code is described as well. We show the score for both approaches for both correct and incorrect Python code, displaying \metric's ability to differentiate between them compared to CodeBERTScore.}
  \label{fig:front-page}
\end{figure*}
\begin{abstract}
AI-based code generation is increasingly prevalent, with GitHub Copilot estimated to generate 46\% of the code on GitHub. Accurately evaluating how well generated code aligns with developer intent remains a critical challenge. Traditional evaluation methods, such as unit tests, are often unscalable and costly. Syntactic similarity metrics (e.g., BLEU, ROUGE) fail to capture code functionality, and metrics like \codebertscore require reference code, which is not always available. To address the gap in reference-free evaluation, with few alternatives such as \icescore, this paper introduces \metric, a novel reference-free metric. \metric uses Contrastive Learning to generate meaningful embeddings for code and natural language task descriptions, enabling similarity scoring that reflects how well generated code implements the task. We show that \metric achieves stronger correlations with functional correctness and human preference than existing metrics across multiple programming languages.
\end{abstract}

\section{Introduction}\label{sec:intro}

The integration of large language models (LLMs) into modern software development has revolutionized the field through automated code generation. This field, often called neural natural-language-to-code (NL2Code), has advanced rapidly with a wave of models, from specialized open-source tools to powerful proprietary systems \cite{allal2023santacoder, zhou2022docprompting, fried2022incoder, codellama, qwen2.5-coder,wizardcoder, Mellum-4b-base, ibm-granite, openai_chatgpt_2025, google_gemini_2025}. These LLMs are revolutionizing software development, with code generation tools like GitHub Copilot now responsible for approximately 46\% of code on GitHub \citep{gao2024copilot}, highlighting the critical need for effective evaluation methods. Although executing generated code against unit tests is considered the gold standard for assessing functional correctness, it often proves impractical. Comprehensive test creation requires significant manual effort, especially in niche languages like Verilog and COBOL, where community-driven testing is limited and automated tools may be unavailable. Proprietary codebases further complicate evaluation when documentation remains locked within companies.

Execution-based methods for code validation evaluate the functional correctness of generated code using testing tools such as unit tests. However, these methods might overlook valuable code snippets that contain minor syntactic errors yet still capture essential logic. For software engineers who can easily fix minor bugs, these functionally sound snippets are highly valuable. This highlights the need for evaluation metrics that prioritize functional correctness over strict syntactic adherence, focusing on a code's ability to perform intended tasks despite imperfections.

Syntactic similarity metrics such as BLEU \citep{papineni2002bleu} and ROUGE \citep{chin2004rouge} offer computationally efficient alternatives but fail to capture the semantic meaning of code, being easily influenced by superficial variations in formatting or identifier names. Some natural language metrics have been enhanced to incorporate code-specific elements like Abstract Syntax Trees (ASTs) and Program Dependency Graphs (PDGs), including RUBY \citep{tran2019doesruby} and CodeBLEU \citep{ren2020codebleu}. Despite these improvements, such metrics often lack correlation with human evaluations of code quality \citep{evtikhiev2023out}.

Recent approaches, such as CodeBERTScore \citep{zhou2023codebertscore}, aim to improve traditional metrics by leveraging pre-trained language models to evaluate the similarity of generated code against a reference implementation---a ground-truth solution known to successfully solve the task. While effective, these approaches fundamentally depend on the availability of reference code, which prevents their use in novel code generation tasks where such references are unavailable or impractical to obtain. To address this limitation, the \icescore metric proposed by \citet{zhuo2023ice} prompts a large language model (LLM) like GPT to evaluate generated code quality directly, without relying on test cases or reference solutions. The LLM assesses various aspects such as correctness and usefulness, making \icescore a flexible, reference-free metric, though this advantage comes at the cost of increased latency and computational resources.
A different approach is taken by \codescore \citep{dong2025codescore}, which trains an LLM to predict functional correctness by estimating how many test cases a snippet passes and whether it runs without errors. \codescore supports three variants: natural language only, reference code only, or both. The two variants involving reference code require the reference implementation, which can be restrictive and costly. Moreover, all three variants require precomputed percentages of passing test cases, further limiting applicability.

To overcome these challenges, we introduce \metric: \textbf{M}etric for \textbf{A}ssessing \textbf{T}ask and \textbf{C}ode \textbf{H}armony, a novel \textit{Contrastive Learning}-based approach to code evaluation that does not require reference code or execution. By learning meaningful embeddings for both code and natural language task descriptions, \metric provides a nuanced assessment of code that captures both functional and semantic correctness. We show that \metric correlates strongly with human judgments of code quality and significantly outperforms existing metrics, particularly in scenarios where traditional evaluation methods are impractical. In \Cref{fig:front-page}, we present examples comparing \metric with \codebertscore, illustrating how similarity to task descriptions helps \metric avoid inherent biases of reference-based methods while yielding scores that reflect syntactic and logical correctness. Additional examples are provided in \cref{appndx:more-examples}.
\paragraph{Contributions:}
In summary, this paper makes the following contributions:

\begin{itemize}
\item We introduce \metric, a new metric for code evaluation based on \textit{Contrastive Learning}.
\item Our metric is designed to be practical and easy to use, requiring only a natural language description of the desired task, without the need for unit tests, existing APIs, or reference code.
\item We demonstrate that our metric correlates well with human evaluations and functional correctness, outperforming existing metric over several popular programming languages.
\end{itemize}

The remainder of this paper is structured as follows: \Cref{sec:related-work} provides a detailed overview of related work in code evaluation. \Cref{sec:setup} formulates the specific problem addressed in this paper. \Cref{sec:methods} describes the new method of \metric. \Cref{sec:exp-setup} and \Cref{sec:experiments} present the experimental setup and benchmark results respectively. \Cref{sec:analysis} presents performance comparisons for different architecture choices. Finally, \Cref{sec:conclusion} concludes the paper and discusses future directions, including a limitations discussion and potential areas for future research. 

\section{Related Work}\label{sec:related-work}

\paragraph{Natural Language Metrics:} Early code evaluation approaches have relied on syntactic similarity metrics adapted from natural language processing, such as BLEU \citep{papineni2002bleu}, which measures the overlap of n-grams between generated and reference text. CrystalBLEU \citep{eghbali2022crystalbleu} modifies BLEU by excluding references that significantly overlap with the input, aiming to provide a more accurate assessment of diverse text generation quality. Other metrics include ROUGE \citep{chin2004rouge}, which evaluates recall by comparing n-gram overlap; METEOR \citep{banerjee2005meteor}, which considers synonyms and stemming; and ChrF \citep{popovic2015chrf}, which focuses on character n-grams. While these metrics compare generated code to reference implementations known to fulfill the intended tasks, they primarily assess syntactic similarity and fail to capture the functional aspects of the code, such as functional correctness.

\paragraph{Adapted Natural Language Metrics for Code:} To address these shortcomings, some metrics have been enhanced to include code-specific elements, such as Abstract Syntax Trees (ASTs), which represent the hierarchical structure of code; Program Dependency Graphs (PDGs), which illustrate the dependencies between program components; and Data Flow Graphs, which depict the flow of data within a program. Examples of such enhanced metrics include RUBY \citep{tran2019doesruby} and CodeBLEU \cite{ren2020codebleu}. CodeBLEU \cite{ren2020codebleu} enhances traditional BLEU by incorporating program-specific features, evaluating similarity through $n$-gram matches, weighted $n$-gram matches, AST matches, and Data Flow matches. Similarly, RUBY, proposed by \cite {tran2019doesruby} is a similarity metric that compares the PGDs of the generated and reference codes; if a PDG cannot be constructed, it falls back to comparing ASTs, and if an AST cannot be constructed, it uses weighted string edit distance between the tokenized reference and generated code. According to \citet{evtikhiev2023out} 's study, despite these adaptations, however, these metrics often lack correlation with human evaluations of code quality.



\paragraph{Advanced Evaluation Metrics for Code Generation:} \codebertscore \citep{zhou2023codebertscore} adapts BERTScore for code by leveraging embeddings from pre-trained code language models such as CodeBERT to measure similarity between generated and reference code. While it captures semantic overlap more effectively than surface-level metrics, it fundamentally requires reference implementations, and its scores are relative, making them difficult to interpret as an absolute measure of correctness for a single code snippet. A different approach is \codescore \citep{dong2025codescore}, an LLM-based metric trained to predict functional correctness by estimating how many test cases a snippet passes and whether it runs without errors. It supports multiple input types—natural language only, reference code only, or both, but all formats require precomputed test-case labels, and some require reference code, which can be restrictive and costly. 
Another alternative is \icescore \citep{zhuo2023ice}, a reference-free metric that prompts an LLM to directly evaluate code quality; however, this approach suffers from high latency and computational costs. These lines of work motivate alternative strategies, such as \textit{Contrastive Learning}, which we leverage in our method to learn meaningful embeddings for code and task descriptions.



\paragraph{Contrastive Learning} is a self-supervised paradigm that learns representations by contrasting positive and negative examples. The core idea is to bring similar instances closer in embedding space while pushing dissimilar ones apart, making it effective in scenarios with limited labeled data. Key works include \cite{oord2018cpc}, which introduced a self-supervised framework using InfoNCE loss, and \cite{radford2021clip}, which demonstrated CLIP’s effectiveness in a multi-modal setting with image-caption pairs. In the context of code, \cite{jain2020contrastive} emphasized semantic functionality through compiler-based transformations to create embeddings capturing code behavior. These works highlight Contrastive Learning’s potential to model underlying code semantics, which we leverage in our evaluation task.

\section{Problem Formulation}\label{sec:setup}
The input consists of: a natural language task description or instruction \( t \in \mathcal{T} \), where \( \mathcal{T} \) represents the set of all possible task descriptions, a code generation model generates a code snippet \( c \in \mathcal{C} \), where \( \mathcal{C} \) is the set of all possible code snippets, and a ground truth label \( y \) that is assigned to evaluate the quality of the code. This label can take one of the following forms:

\begin{itemize}
    \item \textbf{Binary label} \( y_{\text{bin}} \in \{0, 1\} \): Functional correctness is an example of a binary label, where 1 indicates the code is functionally correct for the given task and 0 indicates an incorrect implementation.
    
    \item \textbf{Continuous label} \( y_{\text{cont}} \in [0, S] \): This label provides a continuous score between 0 and \(S\) representing the code's quality. For example, it can indicate how preferable a code snippet is to human developers. As the code becomes more preferable, the score increases towards \(S\), the maximum possible value.
\end{itemize}

We use $y$ to refer to either type of label, depending on the context.

\paragraph{Goal} Our goal is to define a metric function \( f : \mathcal{T} \times \mathcal{C} \rightarrow [-1, 1] \), where for a given task description \( t \) and a code snippet \( c \), the function \( f(t, c) \) produces a metric score that reflects the quality of the code snippet \( c \) with respect to the task described in \( t \). Ideally, the metric function \( f \) should correlate with human evaluations and functional correctness.

In particular, for two candidate code snippets \( c_1, c_2 \in \mathcal{C} \) implementing the same task \( t \in \mathcal{T} \), we want the metric to indicate that if \( c_1 \) performs the task better and is preferred by human evaluators more than \( c_2 \), then \( c_1 \) should receive a higher metric score. Specifically, we seek a function \( f \) such that \( f(t, c_1) > f(t, c_2) \) when \( c_1 \) is considered better than \( c_2 \) based on its functionality and human preference.


\section{\metric}\label{sec:methods}
This section introduces \metric, our method for code evaluation. We detail its neural architecture in \Cref{subsec:architecture} and the contrastive loss objectives used for optimization in \Cref{subsec:objective}

\subsection{Architecture}\label{subsec:architecture}
\begin{figure*}[t]
\centering
  \includegraphics[width=\textwidth]{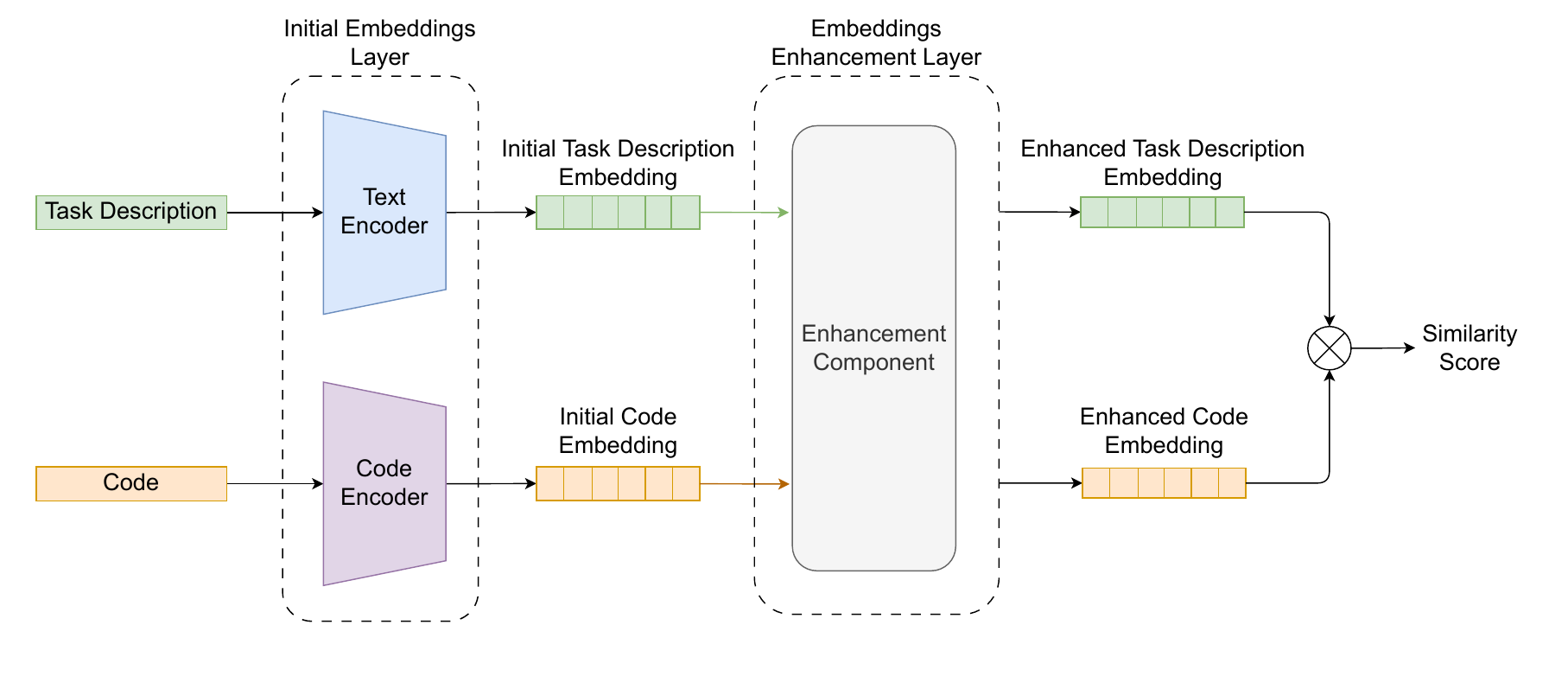}
  \caption{\metric Architecture: This figure illustrates the general architecture of \metric, featuring an Enhanced Embeddings Layer that integrates contextual information from task descriptions and code snippets. Specific implementations, including Linear and Cross-Attention enhancements, are shown in \Cref{fig:enhancement_layers}. The architecture computes similarity between learned embeddings using an appropriate similarity function. such as Cosine Similarity.}
  \label{fig:architecture}
\end{figure*}

\metric generates a similarity score between task descriptions and code snippets by learning embeddings. This process is illustrated in \Cref{fig:architecture}.

%
A key aspect of \metric is its ability to model both functional correctness (successful implementation) and human preferences (developer-favored code) in the similarity space. A close (positive) pair \((t, c)\) represents a task description \(t\) and a corresponding code snippet \(c\) that either successfully implements the task or is highly preferred by developers.


\paragraph{Initial Embeddings}

The initial embedding layer employs a text encoder for the task description $t$ and a code encoder for the code snippet $c$. These encoders can be either trained or kept frozen, an aspect we analyze in \Cref{sec:analysis}. These initial embeddings are then forwarded to the next layer aimed at enhancing and aligning both embeddings within a shared space.

\paragraph{Enhanced Embeddings Layer} 

This layer enhances the initial embeddings by integrating cross-modal information between the task description and code snippet. It is always trained to effectively combine information from both inputs.

Formally, given a task description \( t \in \mathcal{T} \) (in natural language) and its corresponding code snippet \( c \in \mathcal{C} \), we denote the enhanced embeddings for the task description and code snippet as \( \mathbf{e}_{\text{t}} \) and \( \mathbf{e}_{\text{c}} \), respectively.  Both embeddings reside in a \( d \)-dimensional embedding space, where \( d \in \mathbb{Z}^+ \) represents the dimensionality. In general, both $\mathbf{e}_t$ and $\mathbf{e}_c$ depend on both $t$ and $c$, such that:
\begin{equation}\label{eq:en-embeddings}
    \mathbf{e}_t = f_t(t, c) \quad,\quad \mathbf{e}_c = f_c(t, c)
\end{equation}
where $f_t$ and $f_c$ are functions that creates enriched embeddings from $(t, c)$ pairs. These functions aggregate the initial embedding and embedding enhancement blocks as can be seen in \Cref{fig:architecture}.

\begin{figure*}[t]
\centering
  \begin{subfigure}[b]{0.49\textwidth}
    \includegraphics[width=\textwidth]{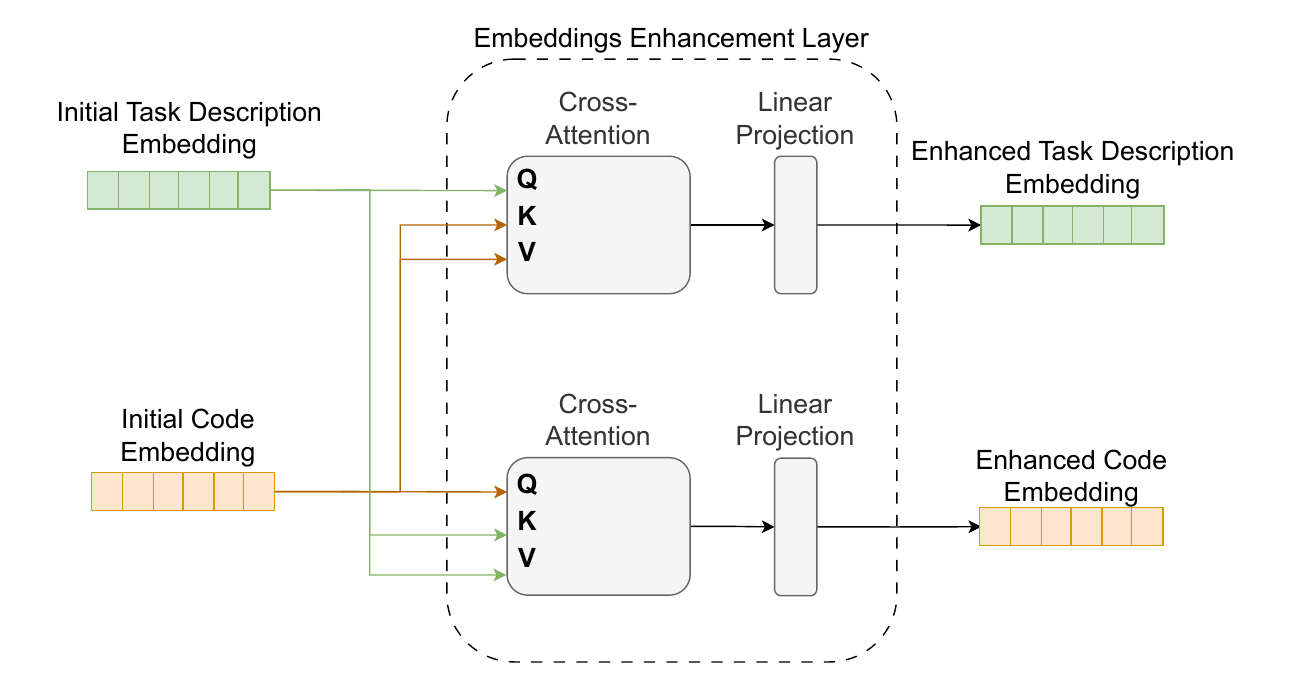}
    \caption{Cross-Attention Layer: which integrates contextual information by using each input as both query and key-value pairs.}
    \label{fig:cross_attention}
  \end{subfigure} \hfill
  \begin{subfigure}[b]{0.49\textwidth}
    \includegraphics[width=\textwidth]{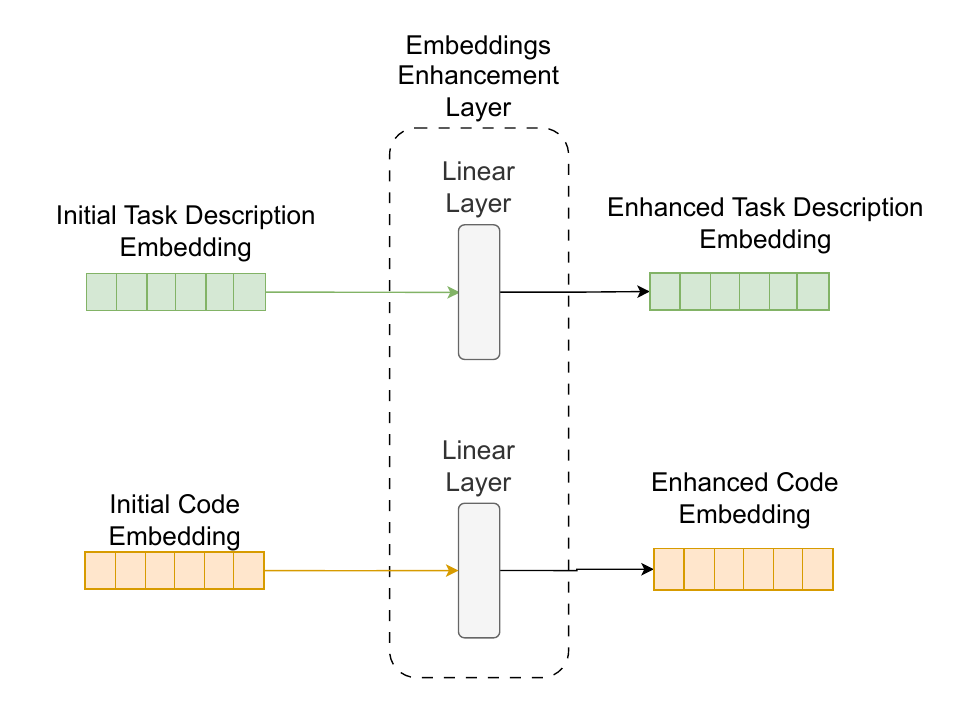}
    \caption{Linear Layer: which applies separate linear transformations to refine features from both the task description and the code snippet.}
    \label{fig:linear}
  \end{subfigure}
  \caption{An illustration of the specific implementations of the Enhanced Embeddings Layer: the Cross-Attention and Linear layers.}
  \label{fig:enhancement_layers}
\end{figure*}

We propose two alternatives for the enhanced embeddings layer:

\begin{itemize}
    \item \textbf{Cross-Attention:} 
    This alternative employs two cross-attention components, one for each input, based on \cite{vaswani2017attention}. For instance, the code component uses code as query and task description as key/value, and vice-versa for the text component. Each component is followed by a linear layer projecting embeddings into a shared space. A detailed illustration can be found in \Cref{fig:cross_attention}.
    
    \item \textbf{Linear:} 
    \textbf{Linear:} This simpler approach applies separate linear transformations to each initial embedding, projecting them into a shared space. Here, $\mathbf{e}_t = f_t(t)$ and $\mathbf{e}_c = f_c(c)$. The shared space alignment is driven by the \textit{Contrastive Learning} objective (\Cref{subsec:objective}). This approach is illustrated in \Cref{fig:linear}.
\end{itemize}

We explore these alternatives further in \Cref{sec:analysis}.

\paragraph{Similarity Score Calculation:} 






We compute the similarity between the enhanced embeddings $\mathbf{e}_t$ and $\mathbf{e}_c$ using \textit{Cosine Similarity}: 

\begin{equation}\label{eq:cos_sim}
\text{cos}(\mathbf{e}_{\text{t}}, \mathbf{e}_{\text{c}}) = \frac{\mathbf{e}_{\text{t}} \cdot \mathbf{e}_{\text{c}}}{\|\mathbf{e}_{\text{t}}\| \, \|\mathbf{e}_{\text{c}}\|}
\end{equation}

The final \metric score, $f(t,c)$, is then given by: 

\begin{equation}\label{eq:ftc}
    f(t, c) = \text{cos}\left(f_t(t,c), f_c(t,c)\right) 
\end{equation}

\subsection{Contrastive Learning  Objective}\label{subsec:objective}


We optimize our architecture using contrastive loss objectives tailored for both binary and continuous label scenarios (as detailed in \Cref{sec:setup}).

Given a task description \( t \in \mathcal{T} \) (in natural language) and a code snippet \( c \in \mathcal{C} \), The following sections \ref{subsec:binary-labels} and \ref{subsec:cont-labels} outline the loss functions for both binary and continuous labels.

\subsubsection{Binary Labels}\label{subsec:binary-labels}

For a binary label $y_{bin} \in \{0, 1\}$ (e.g. indicating whether the code snippet successfully implements the task), we optimize the following loss function for binary loss $\mathcal{L}_{bin}$:

{
\small
\begin{align}
\mathcal{L}_{bin}(t, c, y) =
    &\begin{cases}
    1 - f(t,c) & , y = 1 \\
    \max(0, f(t,c) - m) & , y = 0
    \end{cases}
\end{align}
}

Here, $m$ is a margin that is constrained within the range of $[-1, 1]$, defining the threshold distinguishing between similar and dissimilar pairs, $f(t, c)$ is computed via \Cref{eq:ftc}, and $y \equiv y_{bin}$ for brevity.

\subsubsection{Continuous Labels}\label{subsec:cont-labels}
For a continuous label on a scale  $y_{cont} \in [0, S]$, with $S\in\mathbb{R}^{+}$ for example, a score reflecting human preference regarding the usefulness of the code snippet for the given task, we optimize the following loss function for continuous loss $\mathcal{L}_{cont}$:

\begin{equation}
    \mathcal{L}_{\text{cont}}(t, c, y) = \text{MSE}\left(sim(t, c) - \frac{y}{S}\right)
\end{equation}
Here, $y \equiv y_{cont}$ for brevity, $f(t, c)$ is computed via \Cref{eq:ftc}, and \( sim(t, c)\) is defined as follows:
\begin{equation}
    sim(t, c) \triangleq \frac{1 + f(t, c)}{2}
\end{equation}

\section{Experimental Setup}\label{sec:exp-setup}

\begin{table*}[h]
\centering
\resizebox{\linewidth}{!}{
\begin{tabular}{l|ccc|ccc|ccc|ccc}
\hline
\textbf{Metric} & \multicolumn{3}{c|}{\textbf{Java}} & \multicolumn{3}{c|}{\textbf{Python}} & \multicolumn{3}{c|}{\textbf{JavaScript}} & \multicolumn{3}{c}{\textbf{C++}} \\
 & $\tau$ & $r_s$ & $r_p$ & $\tau$ & $r_s$ & $r_p$ & $\tau$ & $r_s$ & $r_p$  & $\tau$ & $r_s$ & $r_p$\\
\hline
BLEU & .460 & .301 & .291 &	.361 &	.334 &	.274 &	.219 &	.255 &	.242 & .140	& .215	& .131\\
CodeBLEU & .492	&.308	&.318	&.388	&.315	&.323	&.238	&.267	&.296	&.202	&.158	&.157\\
ROUGE-1 & .481	&.341	&.356	&.390	&.334	&.343	&.238	&.276	&.309	&.244	&.319	&.333\\
ROUGE-2 & .436	&.278	&.308	&.365	&.307	&.303	&.199	&.233	&.281	&.221	&.270	&.293\\
ROUGE-L & .464	&.343	&.360	&.382	&.352	&.356	&.207	&.264	&.303	&.245	&.323	&.336\\
METEOR & .511	&.343	&.356	&.426	&.400	&.408	&.257	&.314	&.337	&.204	&.210	&.219\\
chrF &.527	&.346	&.369	&.439	&.385	&.393	&.316	&.344	&.376	&.319	&.331	&.348\\
\codebertscore & .547	&.403	&.406	&.464	&.418	&.388	&.331	&.389	&.327	&.331	&.390	&.379\\
\icescore &.616	&.504	&.499	&.341	&.310	&.315	&\textbf{.540}	&.437	&.435	&\textbf{.509}	&.485	&.486\\
\hline
\textbf{\metric(Base(T), CA)} & .576	&.684	&.675	&.613	&.673	&\textbf{.688}
&.438	&.602	&.593	&.395	&.639	&\textbf{.680}
\\
\textbf{\metric(LS(F), Linear)} & \textbf{.673}	&\textbf{.701}	&\textbf{.700}	&\textbf{.668}	&\textbf{.701}	&.672	&.439	&\textbf{.630}	&\textbf{.626}	&\textbf.494	&\textbf{.688}	&.674\\
\hline
\end{tabular}}
\caption{Correlation of various metrics to functional correctness ($\tau$ = Kendall, $r_s$ = Spearman, and $r_p$ = Pearson) across programming languages using the HumanEval dataset. The best values in each column are bolded. Standard deviation are reported in \Cref{tab:std-humaneval}}
\label{tab:correlations-humaneval}
\end{table*}

\paragraph{Baselines} 
Our main comparisons are with \textbf{\codebertscore} \citep{zhou2023codebertscore}, which enhances code evaluation by incorporating natural language input alongside the generated code and using pre-trained models to assess consistency between them, and \textbf{\icescore} \citep{zhuo2023ice}, which offers a reference-free alternative by employing a large language model (LLM) to evaluate code quality. We further consider \textbf{CodeScore} \citep{dong2025codescore}, an LLM-based metric trained to predict the pass rate of generated code. CodeScore has three reported variants: \emph{(g+n)}, \emph{(g+r)}, and \emph{(g+r+n)}, which differ in the combination of generated code, reference code, and natural language descriptions used as inputs; note two of the three variants require reference code. We adopt this notation when presenting results. Additionally, we compare our metric with established evaluation methods such as \textbf{BLEU}, \textbf{ROUGE}, \textbf{METEOR}, \textbf{chrF}, and \textbf{CodeBLEU}, all of which serve as baselines in our experiments.

\paragraph{MATCH Variants}  
In our experiments, we explore several versions of \metric, defined as \metric(CodeEncoder(X), E). Here, \textbf{CodeEncoder} can be either a base encoder (\textbf{Base}) or a language-specific pre-trained encoder (\textbf{LS}), \(X\) indicates whether the encoder is frozen (\(F\)) or trainable (\(T\)), and \(E\) denotes the enhancement layer type, either cross-attention (\(CA\)) or linear (\(Linear\)). These variants allow us to assess the impact of encoder choice and enhancement architecture on performance.

\paragraph{Correlation metrics} To evaluate our metric, we use three correlation metrics: Kendall's Tau ($\tau$) \cite{kendall1938new}, Pearson's correlation coefficient ($r_p$) \cite{cohen2009pearson}, and Spearman's rank correlation coefficient ($r_s$) \cite{spearman1904}. These metrics follow best practices in natural language evaluation and previous code evaluation studies \cite{zhou2023codebertscore, zhuo2023ice, dong2025codescore}, allowing us to quantify how well each metric’s scores align with reference values.

\section{Experiments}\label{sec:experiments}

In this section, we evaluate the utility of \metric by examining its alignment with different types of labels. The first two experiments assess the correlation of \metric with functional correctness: one using binary correctness labels (\Cref{subsec:functional-experiment}) and the other using pass-rate labels derived from execution tests (\Cref{subsec:passrate-experiment}). The third experiment evaluates the correlation with human preferences for generated code (\Cref{subsec:human-experiment}).

\subsection{Functional Correctness}\label{subsec:functional-experiment}
\paragraph{Dataset}
To evaluate alignment with functional correctness, we use the HumanEval dataset \cite{chen2021evaluating}, which provides natural language descriptions of programming tasks, hand-crafted input-output test cases, and human-written reference solutions. Originally developed for Python, HumanEval has been translated into 18 programming languages by \cite{cassano2022multipl}, and includes predictions generated by the Codex model (code-davinci-002) along with their associated binary functional correctness labels, where each label indicates whether the generated code passes all test cases (1) or not (0). In our experiments, we focus on four widely-used languages: Java, C++, Python, and JavaScript, as highlighted in \cite{zhou2023codebertscore}.

\paragraph{Training}
The task-code pairs for training our model were directly sourced from HumanEval, leveraging its binary functional correctness labels as our optimization target (\( \mathcal{L}_{\text{bin}} \), see \Cref{subsec:binary-labels}). To ensure robust evaluation, we conducted five independent experiments, each using a different random split of the complete dataset into training, validation, and test sets. For each experiment, a separate model was trained using only the training and validation data, then evaluated on the held-out test set to prevent data leakage. Final performance is reported as the mean and standard deviation across all five experiments. Correlation metrics reported in \Cref{tab:correlations-humaneval} are averaged across five splits, with standard deviations provided in \Cref{tab:std-humaneval}. Additional information about the dataset and it attributes, as well as further details on our model’s architecture, training hyperparameters, and implementation, are provided in \Cref{appendix:experimental-detail}.

\paragraph{Results}
For correlation with functional correctness, we report the average correlation scores (Pearson, Spearman, and Kendall’s Tau) across the five test splits, as shown in \Cref{tab:correlations-humaneval}. The corresponding standard deviations are provided in \Cref{tab:std-humaneval}. 

We evaluate two variations of \metric: \textbf{\metric(Base(T), CA)}, which uses a trained CodeBERT-base encoder with a cross-attention enhancement layer, and \textbf{\metric(LS(F), Linear)}, which employs frozen language-specific encoders from \citet{zhou2023codebertscore} with a linear projection layer. Our results show that \metric achieves the highest or comparable Kendall’s Tau correlation with functional correctness across all four programming languages, with \icescore slightly outperforming for JavaScript. Moreover, \metric consistently surpasses all baseline methods in both Spearman and Pearson correlation. Overall, \metric demonstrates stronger alignment with functional correctness than existing evaluation metrics.

\subsection{Functional Correctness with Pass Rate Labels}\label{subsec:passrate-experiment}

\paragraph{Dataset} 
To complement our binary functional correctness experiments, we use MBPP-Eval \citep{dong2025codescore}, a large-scale dataset for functional correctness evaluation. Each task consists of a natural language description, a reference solution, and approximately 24 generated code candidates from various LLMs. Each task also includes around 100 automatically constructed test cases, enabling computation of a continuous pass-rate label (PassRatio) for each code snippet. We replicate the experimental setup from \citet{dong2025codescore}, training on the provided training and development splits and reporting results on the test set. We additionally include \textbf{CodeScore} \citep{dong2025codescore} in this experiment, since MBPP-Eval provides the pass-rate labels required for its training. The dataset contains 15{,}679 training examples and 3{,}000 examples each for development and test sets. The average PassRatio across splits is approximately 0.28, underscoring the difficulty of this benchmark.

\paragraph{Training} 
For this experiment, we optimize our models using $\mathcal{L}_{\text{cont}}$, as defined in \Cref{subsec:cont-labels}, which leverages the continuous PassRatio labels provided by MBPP-Eval. To ensure robustness, we train and evaluate across five random seeds and report the average results. The training and validation sets are used exclusively for optimization and model selection, while the test set is reserved for final evaluation to prevent data leakage. Further details on the model architecture, hyperparameters, and implementation are provided in \Cref{appendix:experimental-detail}.


\paragraph{Results}
\Cref{tab:passrate_results} reports results on the MBPP-Eval benchmark. CodeScore supports three variants: \emph{g+n} (generated code + natural language), \emph{g+r} (generated code + reference code), and \emph{g+r+n} (all three inputs). Since our method relies only on the natural language description of the task, we report the \emph{g+n} variant in the table as the most directly comparable baseline.

Focusing on this fair comparison, \metric achieves higher Kendall–Tau and Spearman correlations, and a comparable Pearson correlation, relative to CodeScore (g+n). This shows that \metric aligns more closely with functional correctness without relying on reference code. CodeScore (g+r) and (g+r+n) attain slightly higher correlations, but both require reference implementations, making them less applicable in scenarios where such references are unavailable.

Across all other baselines, \metric consistently attains the highest correlations for all metrics.

\begin{table}[h]
\centering
\resizebox{\linewidth}{!}{
\begin{tabular}{l|ccc}
\hline
\textbf{Metric} & \multicolumn{3}{c}{\textbf{Python}} \\
 & $\tau$ & $r_s$ & $r_p$ \\
\hline
BLEU & .132 & .188 & .128\\
CodeBLEU & .150 & .213 & .189\\
ROUGE-1 & .268 & .375 & .298\\
ROUGE-2 & .226 & .317 & .261\\
ROUGE-L & .250 & .351 & .289\\
METEOR & .183 & .259 & .223\\
chrF & .180 & .254 & .256\\
\codebertscore & .240 & .338 & .302\\
\icescore & .275 & .330 & .322\\
CodeScore (g+n) & .341 & .488 & \textbf{.577}\\
\hline
\textbf{\metric(LS(F), Linear)} & \textbf{.374} & \textbf{.516} & \textbf{.574}\\
\hline
\end{tabular}}
\caption{Correlation of various metrics to functional correctness ($\tau$ = Kendall, $r_s$ = Spearman, and $r_p$ = Pearson) across on MBPP-Eval dataset. The best values in each column are bolded.}
\label{tab:passrate_results}
\end{table}

\subsection{Human Preference}\label{subsec:human-experiment}

\paragraph{Dataset}
To evaluate the correlation between \metric and human preferences, we use human annotations from \citet{evtikhiev2023out} for the CoNaLa benchmark \cite{yin2018learning}. CoNaLa focuses on Python code generation from natural language descriptions, with data sourced from StackOverflow. For this benchmark, experienced software developers graded code snippets produced by five different models on a scale from 0 to 4, where 0 indicates irrelevance and 4 signifies that the code effectively addresses the problem. These continuous scores serve as human preference labels.

\paragraph{Training}
For this experiment, \metric is trained and evaluated on CoNaLa using the human preference labels. Similar to the functional correctness experiment (\Cref{subsec:functional-experiment}), we conducted five independent experiments using different random splits of the dataset into training, validation, and test sets. In each experiment, the model was trained on the training and validation data and evaluated on the held-out test set to prevent data leakage. Given the continuous nature of the labels, we employ \( \mathcal{L}_{\text{cont}} \), as defined in \Cref{subsec:cont-labels}, as our optimization objective. Additional details on the dataset, its attributes, and our model’s architecture, training hyperparameters, and implementation are provided in \Cref{appendix:experimental-detail}.

\paragraph{Results}
For correlation with human preferences, we report the average metrics (Pearson, Spearman, and Kendall-Tau) across the five test splits, as shown in \Cref{tab:correlaltion-conala}. The standard deviation across splits is provided in \Cref{tab:std-conala}. We find that \textbf{\metric(Base(T), CA)} achieves comparable Kendall-Tau correlation to \codebertscore, while outperforming all baselines in both Spearman and Pearson correlations. These results indicate that \metric aligns more strongly with human judgments overall.

\begin{table}[h]
\centering
\resizebox{\linewidth}{!}{
\begin{tabular}{l|ccc}
\hline
\textbf{Metric} & \multicolumn{3}{c}{\textbf{Python}} \\
 & $\tau$ & $r_s$ & $r_p$ \\
\hline
BLEU & .148 & .272& .286\\
CodeBLEU & .256 & .374 & .424\\
ROUGE-1 & .505 &.633 &.638\\
ROUGE-2 & .357 &.525 &.549\\
ROUGE-L & .488 &.617 &.627\\
METEOR & .168	&.272 & .330\\
chrF &.507 &.622& .626\\
\codebertscore &\textbf{.577} &.662	&.660\\
\icescore &.311 &.561	&.636\\
\hline
\textbf{\metric (Base(T),CA)} & .568	&\textbf{.721}	&\textbf{.741}\\
\hline
\end{tabular}}
\caption{Correlation to functional correctness of various metrics ($\tau$ = Kendall, $r_s$ = Spearman, and $r_p$ = Pearson) on CoNaLa Dataset. Best values per column are bolded. Standard deviation are reported in \Cref{tab:std-conala}}\label{tab:correlaltion-conala}
\end{table}

\section{Analysis}\label{sec:analysis}

\begin{table*}[h]
\centering
\resizebox{\linewidth}{!}{
\begin{tabular}{l|ccc|ccc|ccc|ccc|ccc}
\hline
\textbf{Metric} & \multicolumn{3}{c|}{\textbf{Java}} & \multicolumn{3}{c|}{\textbf{Python}} & \multicolumn{3}{c|}{\textbf{JavaScript}} & \multicolumn{3}{c|}{\textbf{C++}}  & \multicolumn{3}{c}{\textbf{CoNala}} \\
 & $\tau$ & $r_s$ & $r_p$ & $\tau$ & $r_s$ & $r_p$ & $\tau$ & $r_s$ & $r_p$  & $\tau$ & $r_s$ & $r_p$ & $\tau$ & $r_s$ & $r_p$\\
\hline
\metric(LS(F),Linear) & \textbf{.673}	&\textbf{.701}	&\textbf{.700}	&\textbf{.668}	&\textbf{.701}	&.672	&.439	&\textbf{.630}	&\textbf{.626}	&\textbf{.494}	&\textbf{.688}	&.674 &.503	&.679 &.712\\
\metric(LS(F),CA) &.635	&.692	&.663	&.604	&.687	&.645	&.385	&.603	&.566	&.450	&.678	&.665 &.442	&.642	&.671\\
\metric(Base(T),CA) &.576	&.684	&.675	&.613	&.673	&\textbf{.688}
	&.438	&.602	&.593	&.395	&.639	&\textbf{.680} &\textbf{.568}	&\textbf{.721}	&\textbf{.741}\\
\metric (Base(T),Linear) &.510	&.581	&.539	&.566	&.657	&.661	&\textbf{.464}	&.624	&.613	&.305	&.636	&.647 &\textbf{.560}	&\textbf{.726}	&\textbf{.744}
\\

\hline
\end{tabular}}
\caption{Correlation to functional correctness of different variations of \metric ($\tau$ = Kendall, $r_s$ = Spearman, and $r_p$ = Pearson) across programming languages from HumanEval and correlation to human preferences on CoNaLa Datasets. Best values per column are bolded. Standard devaitions are presented in \Cref{tab:analysis-std}.}
\label{tab:analysis}
\end{table*}

This section presents an analysis of the different components within our architecture and examines the trade-offs associated with various design choices, as shown in \Cref{tab:analysis}. We address the following aspects:

\paragraph{Impact of Language-Specific Code Encoders vs. a Base Encoder}
\Cref{tab:analysis} shows that using a language-specific code encoder, when available, yields improved correlations with functional correctness. However, a base code encoder tends to achieve better correlations with human preference. 

\paragraph{Impact of Training the Code Encoder vs. Freezing It}
\Cref{tab:analysis} indicates that training a base code encoder achieves reasonable results when a language-specific encoder is not available. When language-specific encoders are available, \metric generally achieves better results. For the CoNaLa dataset, training the base code encoder consistently results in higher correlations with human preference. However, for HumanEval, freezing a language-specific encoder generally yields better performance.

\paragraph{Impact of a Cross-Attention Enhancement Layer vs. a Linear Layer}
\Cref{tab:analysis} suggests that when a language-specific code encoder is available, a linear enhancement layer is sufficient and achieves strong correlations with functional correctness. Conversely, in the absence of a language-specific encoder, a cross-attention layer can compensate and yield good correlations with functional correctness. Furthermore, a linear enhancement layer consistently achieves the best correlations with human preference.

\textbf{Architecture Choice - Summary}
\metric using language-specific encoders tends to correlate better with functional correctness. When these encoders are available, a linear enhancement layer (Lang(F), Linear) is often sufficient and resource-efficient. Alternatively, \metric with a base code encoder and cross-attention (Base, CA) provides a good trade-off: it can perform reasonably well while requiring less data and computational resources than training or using language-specific encoders. Notably, when using the same code encoder, \metric consistently shows a stronger correlation with human preference when a linear enhancement layer is used compared to cross-attention. Overall, \metric provides a flexible framework for code evaluation, allowing users to select architectures that balance performance, data needs, and resource availability. The continued investigation of different architectures within \metric presents a promising and interesting avenue for future work.

\section{Conclusion}\label{sec:conclusion}


We introduced \metric, a practical, reference-free metric for evaluating code implementations against natural language descriptions. Addressing the critical challenge of assessing code quality without requiring reference implementations or extensive test cases, \metric leverages contrastive learning to embed code and text into a shared semantic space. Our method first independently encodes code and natural language, then enriches their representations through Cross-modal Interaction. To quantify alignment using cosine similarity. Empirical experiments demonstrated \metric's superior performance, achieving significantly stronger correlations with both functional correctness and human preference compared to existing baselines. This positions \metric as a powerful and flexible solution for efficiently assessing code quality, particularly in scenarios where traditional evaluation methods are impractical. 

Looking ahead, \metric can be extended to evaluate a broader spectrum of code quality beyond functional correctness and human preference. This includes non-functional aspects such as efficiency (time and space complexity), readability, maintainability, and security vulnerabilities, enabling a more holistic and practical assessment of generated code. Another promising direction is testing the method’s robustness under imperfect or noisy inputs, which are common in real-world scenarios.

\section*{Limitations}\label{sec:limitations}
\metric is primarily designed with software developers in mind. As such, it tends to assign high scores to code that is syntactically and semantically correct, even if it contains minor typos. While this is desired for developer-facing tools and early-stage prototyping, it may not suffice for evaluating production-level code, where robustness, error handling, and performance are critical. To address these gaps, we recommend supplementing \metric with additional evaluation strategies, such as unit tests for edge cases or compilation checks.

Another consideration is the reliance on model fine-tuning. Our results indicate that the performance of the proposed metric improves when the underlying model is fine-tuned on task-specific data. However, this fine-tuning process can be computationally expensive and may not be practical in all settings. Fortunately, our experiments show that even without fine-tuning, \metric achieves competitive performance, suggesting it can be effectively applied in low-resource or general-use scenarios.

Finally, while our method performs well on established benchmarks that are widely used and well-understood in the community, its effectiveness on entirely new or out-of-distribution tasks remains uncertain. This limitation is not unique to our approach, it is a fundamental challenge shared by all existing and future evaluation metrics aimed at providing universal assessment of code generation and natural language tasks.

\section*{Acknowledgments}
The authors would like to thank Amir Feder for their valuable feedback and advice on this paper. We are also grateful to the anonymous reviewers for their useful discussion and comments.

\bibliography{references}

\newpage
\appendix
\section{Additional Experimental Detail}
\label{appendix:experimental-detail}

Here we provide additional experimental details and full results for our experiment for Correlation to functional correctness and human preference in \Cref{appendix:exp-humaneval} and \Cref{appendix:exp-conala}, respectively.

for both experiments we do the following, we split the data to $5$ splits, and for each split we run all methods, and report the mean correlation for each Method.
For \codebertscore \citep{zhou2023codebertscore} we obtian the scores according to the instruction in  \href{https://github.com/neulab/code-bert-score/tree/main}{https://github.com/neulab/code-bert-score/tree/main}.
For \icescore we run their method following the instructions in 
\href{https://github.com/terryyz/ice-score/tree/main}{https://github.com/terryyz/ice-score/tree/main}. We modify their evaluator by calling the LLM using an interface we have access to. Originally, they use GPT-3.5 (\texttt{GPT-3.5-turbo3}) \footnote{https://platform.openai.com/docs/models/gpt-3-5}
). Due to the fact that we don't have access to this version, we run their method with (\texttt{GPT-4o})\footnote{https://platform.openai.com/docs/models/gpt-4o}.
As for \metric, for the Encoder we use \texttt{
bert-base-uncased} as the text encoder. For the code encoder, we use \texttt{microsoft/codebert-base} for the base encoder, and \texttt{neulab/codebert-{lang}} by \citet{zhou2023codebertscore}, for $\text{lang}\in \{\text{python},\text{java},\text{javascript},\text{cpp}\}$.
\metric's  was implemented using
\texttt{pytorch\_lightning} model as a wrapper model that can receive any model for both encoders. and any model as the enhancement layer.
The \metric models were trained with
\texttt{learning\_rate} $=3e-5$, \texttt{max\_epochs} $=50$, and early stopping with \texttt{patience} $=3$, and a \texttt{batch\_size}=$16$ In addition, we used \texttt{temperature} $=0.07$ for smoothing the similarities in loss calculation.
The final embedding dimensions were \texttt{embedding\_dim},$=768$.

For the Cross-attention enhancements layer, we use one  \texttt{pytorch }\texttt{MultiheadAttention}, with             \texttt{num\_heads}$=8$, \texttt{dropout\_rate}$=0.2$, and \texttt{embed\_dim}$=0.2$, followed by a normalization and linear layers.
with \texttt{in\_features}$=768$, and \texttt{out\_features}$=768$.
As for the Linear enhancement layer, we used a straightforward linear layer with the same parameters.
           
Finally We used 240 GPU-hours on a single A100 GPU.

\paragraph{Baseline Metrics}
In \Cref{tab:baselines-packages} we summarize the used existing packages for each of the baselines metrics we compare \metric to in \Cref{sec:experiments}.
These packages are all licensed under either the MIT or Apache-2.0 license.

\begin{table*}
  \centering
  \begin{tabular}{l|l}
    \hline
    \textbf{Metric}           & \textbf{Package}\\
    \hline
    BLEU           &  \href{https://www.nltk.org/}{nltk}\\
    CodeBLEU       &     \href{https://github.com/k4black/codebleu}{CodeBlEU}\\
    ROUGE          & \href{https://www.nltk.org/}{nltk}\\
    METEOR         & \href{https://www.nltk.org/}{nltk} \\
    chRF          & \href{https://huggingface.co/spaces/evaluate-metric/sacrebleu}{sacrebleu}\\
    \codebertscore &         \href{https://github.com/neulab/code-bert-score/tree/main}{CodeBERTScore}    \\
    \icescore      &  \href{https://github.com/terryyz/ice-score}{ICE-Score}                      \\
    \hline
  \end{tabular}
  \caption{\label{tab:baselines-packages} Packages we used for the baselines we compare our method to in the experiments.
  }
\end{table*}

\subsection{Functional Correctness Experiment}\label{appendix:exp-humaneval}

\paragraph{HumanEval Dataset}
The HumanEval benchmark \citep{chen2021evaluating} is a widely used dataset for evaluating code generation models, consisting of 164 Python programming problems, each with a natural language goal in English, human-written input-output test cases (averaging $7.7$ per problem), and a human-written reference solution. Each example aims to evaluate the model on functional correctness. While the original HumanEval is in Python, an extension by \citet{cassano2022multipl} translated the problems to $18$ other languages, including Java, C++, and JavaScript, facilitating evaluation across diverse programming paradigms. This translated version also includes predictions from \texttt{code-davinci-002} along with their corresponding functional correctness scores.  Utilizing data from the HumanEval-X dataset \citet{zheng2023codegeex} to provide reference solutions in the translated languages, in work we focuses on evaluating code generation performance in Java, C++, Python, and JavaScript.

\begin{table*}[h]
\centering
\resizebox{\linewidth}{!}{
\begin{tabular}{l|ccc|ccc|ccc|ccc}
\hline
\textbf{Metric} & \multicolumn{3}{c|}{\textbf{Java}} & \multicolumn{3}{c|}{\textbf{Python}} & \multicolumn{3}{c|}{\textbf{JavaScript}} & \multicolumn{3}{c}{\textbf{C++}} \\
 & $\tau$ & $r_s$ & $r_p$ & $\tau$ & $r_s$ & $r_p$ & $\tau$ & $r_s$ & $r_p$  & $\tau$ & $r_s$ & $r_p$\\
\hline
BLEU & .460$\pm$.017 & .301$\pm$.006 & .291$\pm$.012 & .361$\pm$.036 & .334$\pm$.010 & .274$\pm$.015 & .219$\pm$.044 & .255$\pm$.047 & .242$\pm$.039 & .140$\pm$.019 & .215$\pm$.023 & .131$\pm$.015\\
CodeBLEU & .492$\pm$.014 & .308$\pm$.011 & .318$\pm$.016 & .388$\pm$.027 & .315$\pm$.011 & .323$\pm$.018 & .238$\pm$.052 & .267$\pm$.024 & .296$\pm$.031 & .202$\pm$.012 & .158$\pm$.008 & .157$\pm$.006\\
ROUGE-1 & .481$\pm$.009 & .341$\pm$.007 & .356$\pm$.005 & .390$\pm$.044 & .334$\pm$.019 & .343$\pm$.017 & .238$\pm$.057 & .276$\pm$.029 & .309$\pm$.030 & .244$\pm$.014 & .319$\pm$.011 & .333$\pm$.010\\
ROUGE-2 & .436$\pm$.014 & .278$\pm$.010 & .308$\pm$.009 & .365$\pm$.035 & .307$\pm$.014 & .303$\pm$.015 & .199$\pm$.043 & .233$\pm$.041 & .281$\pm$.040 & .221$\pm$.018 & .270$\pm$.009 & .293$\pm$.011\\
ROUGE-L & .464$\pm$.015 & .343$\pm$.006 & .360$\pm$.006 & .382$\pm$.037 & .352$\pm$.021 & .356$\pm$.018 & .207$\pm$.069 & .264$\pm$.033 & .303$\pm$.036 & .245$\pm$.016 & .323$\pm$.013 & .336$\pm$.013\\
METEOR & .511$\pm$.022 & .343$\pm$.007 & .356$\pm$.011 & .426$\pm$.036 & .400$\pm$.013 & .408$\pm$.015 & .257$\pm$.048 & .314$\pm$.040 & .337$\pm$.040 & .204$\pm$.030 & .210$\pm$.014 & .219$\pm$.010\\
chrF & .527$\pm$.009 & .346$\pm$.014 & .369$\pm$.012 & .439$\pm$.024 & .385$\pm$.012 & .393$\pm$.016 & .316$\pm$.031 & .344$\pm$.032 & .376$\pm$.030 & .319$\pm$.014 & .331$\pm$.010 & .348$\pm$.001\\
\codebertscore & .547$\pm$.007 & .403$\pm$.009 & .406$\pm$.008 & .464$\pm$.028 & .418$\pm$.015 & .388$\pm$.012 & .331$\pm$.066 & .389$\pm$.025 & .327$\pm$.017 & .331$\pm$.017 & .390$\pm$.013 & .379$\pm$.010\\
\icescore & .616$\pm$.021 & .504$\pm$.013 & .499$\pm$.012 & .341$\pm$.015 & .310$\pm$.013 & .315$\pm$.013 & \textbf{.540$\pm$.072} & .437$\pm$.033 & .435$\pm$.031 & \textbf{.509$\pm$.012} & .485$\pm$.015 & .486$\pm$.015\\
\hline
\textbf{\metric(Base(T), CA)} & .576$\pm$.046 & .684$\pm$.012 & .675$\pm$.017 & .613$\pm$.026 & .673$\pm$.013 & \textbf{.688$\pm$.016} & .438$\pm$.074 & .602$\pm$.041 & .593$\pm$.047 & .395$\pm$.081 & .639$\pm$.027 & \textbf{.680$\pm$.015}\\
\textbf{\metric(LS(F), Linear)} & \textbf{.673$\pm$.025} & \textbf{.701$\pm$.013} & \textbf{.700$\pm$.017} & \textbf{.668$\pm$.029} & \textbf{.701$\pm$.011} & .672$\pm$.015 & .439$\pm$.107 & \textbf{.630$\pm$.021} & \textbf{.626$\pm$.021} & \textbf{.494$\pm$.015} & \textbf{.688$\pm$.013} & .674$\pm$.012\\
\hline
\end{tabular}}
\caption{Mean and standard deviation of the correlations of various metrics ($\tau$ = Kendall, $r_s$ = Spearman, and $r_p$ = Pearson) across programming languages from HumanEval Dataset. Correlations are presented in \Cref{tab:correlations-humaneval}.}
\label{tab:std-humaneval}.
\end{table*}

\subsection{Human preference Experiment}\label{appendix:exp-conala}
\paragraph{CoNaLa Dataset}
The CoNaLa benchmark \cite{yin2018learning} serves as a dataset for assessing the capability of models to generate Python code from natural language instructions written in English. which focuses on generating Python code from natural language descriptions sourced from StackOverflow. \citet{evtikhiev2023out} provided human annotations for the dataset. Experienced software developers graded the code snippets generated by five models on a scale from zero to four, where zero indicates irrelevance and four signifies that the code effectively addresses the problem.This dataset features a collection of 2,860 code snippets, produced by five different models across 472 distinct examples from CoNaLa. Each generated snippet has been assessed by approximately 4.5 experienced software developers. These human annotations provide a means to correlate automated evaluation metrics with human preferences in code generation quality.

\begin{table}[h]
\centering
\resizebox{\linewidth}{!}{
\begin{tabular}{l|ccc}
\hline
\textbf{Metric} & \multicolumn{3}{c}{\textbf{Python}} \\
 & $\tau$ & $r_s$ & $r_p$ \\
\hline
BLEU & $.148 \pm .061$ & $.272 \pm .010$ & $.286 \pm .022$ \\
CodeBLEU & $.256 \pm .039$ & $.374 \pm .034$ & $.424 \pm .030$ \\
ROUGE-1 & $.505 \pm .049$ & $.633 \pm .036$ & $.638 \pm .030$ \\
ROUGE-2 & $.357 \pm .063$ & $.525 \pm .027$ & $.549 \pm .030$ \\
ROUGE-L & $.488 \pm .060$ & $.617 \pm .036$ & $.627 \pm .027$ \\
METEOR & $.168 \pm .053$ & $.272 \pm .009$ & $.330 \pm .033$ \\
chrF & $.507 \pm .066$ & $.622 \pm .024$ & $.626 \pm .028$ \\
\codebertscore & $\mathbf{.577} \pm .074$ & $.662 \pm .033$ & $.660 \pm .033$ \\
\icescore & $.311 \pm .060$ & $.561 \pm .036$ & $.636 \pm .027$ \\
\hline
\textbf{\metric (Base(T),CA)} & $.568 \pm .065$ & $\mathbf{.721} \pm .027$ & $\mathbf{.741} \pm .021$ \\
\hline
\end{tabular}}
\caption{Mean and standard deviation of the correlations of various metrics ($\tau$ = Kendall, $r_s$ = Spearman, and $r_p$ = Pearson) on CoNaLa Dataset. Correlations are presented in \Cref{tab:correlaltion-conala}.}
\label{tab:std-conala}
\end{table}

\subsection{Human Preference}\label{appndx:human-experiment}
\paragraph{Dataset} To evaluate the correlation between each metric and human preferences, we utilize human annotations from \citet{evtikhiev2023out} for the CoNaLa benchmark \cite{yin2018learning}, which focuses on generating Python code from natural language descriptions sourced from StackOverflow. Experienced software developers graded the code snippets generated by five models on a scale from zero to four, where zero indicates irrelevance and four signifies that the code effectively addresses the problem. 

In this experiment, the dataset contains continuous labels. Therefore, we will use \( \mathcal{L}_{\text{cont}} \), as defined in \Cref{subsec:cont-labels}, as our optimization objective. Further details can be found in \Cref{appendix:exp-conala}.

\subsection{Analysis Standard Deviations}
Standard deviations of the analysis presented in \Cref{tab:analysis} are reported in \Cref{tab:analysis-std}.

\begin{table*}[h]
\centering
\resizebox{\linewidth}{!}{
\begin{tabular}{l|ccc|ccc|ccc|ccc|ccc}
\hline
\textbf{Metric} & \multicolumn{3}{c|}{\textbf{Java}} & \multicolumn{3}{c|}{\textbf{Python}} & \multicolumn{3}{c|}{\textbf{JavaScript}} & \multicolumn{3}{c|}{\textbf{C++}}  & \multicolumn{3}{c}{\textbf{CoNala}} \\
 & $\tau$ & $r_s$ & $r_p$ & $\tau$ & $r_s$ & $r_p$ & $\tau$ & $r_s$ & $r_p$  & $\tau$ & $r_s$ & $r_p$ & $\tau$ & $r_s$ & $r_p$\\
\hline
\metric(LS(F),Linear) & $\mathbf{.673 \pm .025}$ & $\mathbf{.701 \pm .013}$ & $\mathbf{.700 \pm .017}$ & $\mathbf{.668 \pm .029}$ & $\mathbf{.701 \pm .011}$ & $.672 \pm .015$ & $.439 \pm .107$ & $\mathbf{.630 \pm .021}$ & $\mathbf{.626 \pm .021}$ & $\mathbf{.494 \pm .015}$ & $\mathbf{.688 \pm .013}$ & $.674 \pm .012$ & $.503 \pm .059$ & $.679 \pm .016$ & $.712 \pm .015$ \\
\metric(LS(F),CA) & $.635 \pm .020$ & $.692 \pm .006$ & $.663 \pm .021$ & $.604 \pm .028$ & $.687 \pm .010$ & $.645 \pm .016$ & $.385 \pm .099$ & $.603 \pm .020$ & $.566 \pm .032$ & $.450 \pm .032$ & $.678 \pm .009$ & $.665 \pm .007$ & $.442 \pm .104$ & $.642 \pm .014$ & $.671 \pm .022$ \\
\metric(Base(T),CA) & $.576 \pm .046$ & $.684 \pm .012$ & $.675 \pm .017$ & $.613 \pm .026$ & $.673 \pm .013$ & $\mathbf{.688 \pm .016}$ & $.438 \pm .074$ & $.602 \pm .041$ & $.593 \pm .047$ & $.395 \pm .081$ & $.639 \pm .027$ & $\mathbf{.680 \pm .015}$ & $\mathbf{.568 \pm .065}$ & $\mathbf{.721 \pm .027}$ & $\mathbf{.741 \pm .021}$ \\
\metric(Base(T),Linear) & $.510 \pm .119$ & $.581 \pm .131$ & $.539 \pm .193$ & $.566 \pm .042$ & $.657 \pm .022$ & $.661 \pm .035$ & $\mathbf{.464 \pm .118}$ & $.624 \pm .030$ & $.613 \pm .024$ & $.305 \pm .063$ & $.636 \pm .017$ & $.647 \pm .030$ & $.560 \pm .063$ & $\mathbf{.726 \pm .024}$ & $\mathbf{.744 \pm .025}$ \\
\hline
\end{tabular}}
\caption{Mean and standard deviation of the correlation to functional correctness of different variations of \metric ($\tau$ = Kendall, $r_s$ = Spearman, and $r_p$ = Pearson) across programming languages from HumanEval and correlation to human preferences on CoNaLa Datasets. Correlations are presented in \Cref{tab:analysis}.}
\label{tab:analysis-std}
\end{table*}

\clearpage  
\onecolumn  

\section{Examples}\label{appndx:more-examples}

\newcolumntype{Y}{>{\raggedright\arraybackslash}X}

\begin{table}[H]
\centering
\renewcommand{\arraystretch}{1.2}
\scriptsize

\begin{tabularx}{\textwidth}{|X|}
\hline
\textbf{Example 1: MATCH identifies a mostly correct candidate with a minor bug.} \\
The code is mostly correct but contains a minor bug related to case sensitivity in vowel removal, affecting the consonant count. The MATCH score indicates its potential despite this issue. \\
\hline

\end{tabularx}

\vspace{0.5em}

\begin{tabularx}{\textwidth}{|l|l|X|X|c|c|c|}
\hline
\textbf{Language} & \textbf{Aspect} & \textbf{Task Description} & \textbf{Code Candidate} & \textbf{Label} & \textbf{ICE} & \textbf{MATCH} \\
\hline
Python & Functional Correctness &
Given a string s and a natural number n, implement a function that returns all words from string s with exactly n consonants in order. Return empty list if s is empty.
&
\begin{minipage}{\linewidth}
\ttfamily\footnotesize
arr = []\\
for word in s.split():\\
\hspace{1em}s = word.replace("a", "").replace("e", "")\textbackslash\\
\hspace{2em}.replace("i", "").replace("o", "")\textbackslash\\
\hspace{2em}.replace("u", "")\\
\hspace{1em}if len(s) == n:\\
\hspace{2em}arr.append(word)\\
return arr\\
\end{minipage}
& 0 & 0 & 0.774 \\
\hline
\end{tabularx}
\end{table}

\vspace{1em}

\begin{table}[H]
\centering
\renewcommand{\arraystretch}{1.2}
\scriptsize
\begin{tabularx}{\textwidth}{|X|}
\hline
\textbf{Example 2: MATCH shows a clearer separation between wrong and right solutions.}\\
The first candidate fails to correctly implement the task by mis-checking the oddness of the first digit and neglecting the condition of being greater than 10 while the second candidate performs correctly. MATCH score reflects this significant error, while CodeBERTScore gives close scores for both candidates. \\
\hline
\end{tabularx}

\vspace{0.5em}

\begin{tabularx}{\textwidth}{|l|l|X|X|c|c|c|}
\hline
\textbf{Language} & \textbf{Aspect} & \textbf{Task Description} & \textbf{Code Candidate} & \textbf{Label} & \textbf{ICE} & \textbf{MATCH} \\
\hline
\multirow{2}{*}{Java} & 
\multirow{2}{*}{\parbox{1.5cm}{Functional\\Correctness}} & 
\multirow{2}{*}{\parbox{3cm}{Write a function that counts numbers >10 with odd first and last digits.}} &
\parbox{4cm}{%
\ttfamily
\raggedright
return (int) nums.stream()\\
\hspace{0.5em}.filter(i -> Math.abs(i) \% 2 == 1\\
\hspace{1em}\&\& i \% 10 \% 2 == 1\\
\hspace{1em}\&\& i / 10 \% 2 == 1)\\
\hspace{0.5em}.count();
}
& 0 & 0.699 & 0.218 \\
\cline{4-7}
& & &
\parbox{4cm}{%
\ttfamily
\raggedright
return (int) nums.stream()\\
\hspace{0.5em}.filter(num -> num > 10)\\
\hspace{0.5em}.filter(num ->\\
\hspace{1em}Integer.toString(num)\\
\hspace{1.5em}.charAt(0) \% 2 != 0\\
\hspace{1em}\&\& Integer.toString(num)\\
\hspace{1.5em}.charAt(Integer.toString\\
\hspace{2em}(num).length() - 1)\\
\hspace{1.5em}\% 2 != 0\\
\hspace{0.5em}).count();
}
& 1 & 0.763 & 0.714 \\
\hline
\end{tabularx}
\end{table}

\begin{table}[H]
\centering
\renewcommand{\arraystretch}{1.2}
\scriptsize

\begin{tabularx}{\textwidth}{|X|}
\hline
\textbf{Example 3: MATCH has a better understanding of the code semantics.}\\
Both code candidates fail to correctly implement the task. The first candidate is incorrect, while the second candidate sums the binary digits instead of the decimal digits. The MATCH score accurately reflects the errors in both candidates, unlike the ICE-Score, which misclassifies the second candidate as correct.\\
\hline
\end{tabularx}

\vspace{0.5em}

\begin{tabularx}{\textwidth}{|c|c|X|X|c|c|c|c|}
\hline
\textbf{Language} & \textbf{Aspect} & \textbf{Task Description} & \textbf{Code Candidate} & \textbf{Label} & \textbf{CodeBERTScore} & \textbf{ICE} & \textbf{MATCH} \\
\hline
\multirow{2}{*}{Java} &
\multirow{2}{*}{\parbox{0.1\textwidth}{Functional\\Correctness}} &
\multirow{2}{*}{\parbox{\linewidth}{Given a positive integer N, return the total sum of its digits in binary.}} &
\begin{minipage}[t]{\linewidth}
\ttfamily\tiny
return null;
\end{minipage}
& 0 & 0.617 & 0 & 0.125 \\
\cline{4-8}
& & &
\begin{minipage}[t]{\linewidth}
\ttfamily\tiny
String s = Integer.toBinaryString(N);\\
int sum = 0;\\
for (int i = 0; i < s.length(); i++) \{\\
\hspace{1em}sum += Integer.parseInt(\\
\hspace{2em}String.valueOf(s.charAt(i)));\\
\}\\
return Integer.toString(sum);
\end{minipage}
& 0 & 0.896 & 4 & 0.331 \\
\hline
\end{tabularx}
\end{table}

\begin{table}[H]
\centering
\renewcommand{\arraystretch}{1.2}
\scriptsize

\begin{tabularx}{\textwidth}{|X|}
\hline
\textbf{Example 4: MATCH Effectively Differentiating between correct and incorrect code}\\
The first code candidate correctly downloads a file, while the second incorrectly attempts to open a file without downloading it. The MATCH score reflects this, but the CodeBERTScore does not adequately differentiate between the candidates, and the ICE-Score gives both a score of 0.\\
\hline
\end{tabularx}

\vspace{0.5em}

\begin{tabularx}{\textwidth}{|c|c|X|X|c|c|c|c|}
\hline
\textbf{Language} & \textbf{Aspect} & \textbf{Task Description} & \textbf{Code Candidate} & \textbf{Label} & \textbf{CodeBERTScore} & \textbf{ICE} & \textbf{MATCH} \\
\hline
\multirow{2}{*}{Python} &
\multirow{2}{*}{\parbox{0.1\textwidth}{Human\\Preference}} &
\multirow{2}{*}{\parbox{\linewidth}{Download a file "url" over HTTP and save to "mp3.mp3".}} &
\begin{minipage}[t]{\linewidth}
\ttfamily\tiny
\url{import urllib2, os;} \\
\url{urllib2.urlretrieve("http://www.example.com/songs/mp3.mp3", "mp3.mp3")}
\end{minipage}
& 3.75 & 0.945 & 0 & 0.727 \\
\cline{4-8}
& & &
\begin{minipage}[t]{\linewidth}
\ttfamily\tiny
\url{with open("filename.txt", shell=True)}
\end{minipage}
& 0 & 0.641 & 0 & 0.139 \\
\hline
\end{tabularx}
\end{table}


\begin{table}[H]
\centering
\renewcommand{\arraystretch}{1.2}
\scriptsize

\begin{tabularx}{\textwidth}{|X|}
\hline
\textbf{Example 5: MATCH Effectively Differentiating between correct and incorrect code}\\
The first candidate is not useful at all as it only reads the file without erasing its contents. In contrast, the second candidate correctly opens the file in write mode, successfully erasing its contents. This example shows that the MATCH score better differentiates between the candidates than CodeBERTScore.\\
\hline
\end{tabularx}

\vspace{0.5em}

\begin{tabularx}{\textwidth}{|c|c|X|X|c|c|c|}
\hline
\textbf{Language} & \textbf{Aspect} & \textbf{Task Description} & \textbf{Code Candidate} & \textbf{Label} & \textbf{CodeBERTScore} & \textbf{MATCH} \\
\hline
\multirow{2}{*}{Python} &
\multirow{2}{*}{\parbox{0.1\textwidth}{Human\\Preference}} &
\multirow{2}{*}{\parbox{\linewidth}{Erase all the contents of a file filename.}} &
\begin{minipage}[t]{\linewidth}
\ttfamily\tiny
\url{with open("filename.txt") as f:} \\
\url{\#NEWLINE\#} \\
\url{\#INDENT\#f.read()}
\end{minipage}
& 0.667 & 0.78 & -0.55 \\
\cline{4-7}
& & &
\begin{minipage}[t]{\linewidth}
\ttfamily\tiny
\url{open(filename, w).close()}
\end{minipage}
& 4 & 0.984 & 0.919 \\
\hline
\end{tabularx}
\end{table}

\end{document}